\documentclass[11pt]{article}
\usepackage[preprint]{acl}
\usepackage{times}
\usepackage{latexsym}
\usepackage[T1]{fontenc}
\usepackage[utf8]{inputenc}
\usepackage{microtype}
\usepackage{inconsolata}
\usepackage{graphicx}
\usepackage{booktabs}
\usepackage{multirow}
\usepackage{amsmath,amssymb}
\usepackage{bm}
\usepackage{xcolor}
\usepackage{colortbl}
\definecolor{heat}{HTML}{2E9E5B}
\usepackage{listings}
\usepackage[most]{tcolorbox}
\lstdefinestyle{promptstyle}{
  basicstyle=\ttfamily\footnotesize,
  breaklines=true,
  breakatwhitespace=true,
  breakautoindent=false,
  breakindent=0pt,
  columns=fullflexible,
  keepspaces=true,
  showstringspaces=false,
  frame=none, xleftmargin=0pt, xrightmargin=0pt, framesep=0pt,
  aboveskip=0pt,
  belowskip=0pt,
  moredelim=*[is][\color{blue!80!black}]{`}{`},
  literate={~ND~}{{--}}1 {~EL~}{{...}}1 {~NL~}{{$\hookleftarrow$}}1,
}
\tcbset{
  promptbox/.style={
    enhanced,
    breakable=false,
    colback=white,
    colframe=black!55,
    colbacktitle=black,
    coltitle=white,
    boxrule=0.5pt,
    titlerule=0pt,
    arc=2pt,
    left=4pt, right=4pt, top=3pt, bottom=3pt,
    toptitle=2pt, bottomtitle=2pt,
    fonttitle=\ttfamily\bfseries\footnotesize,
  },
  promptboxdata/.style={promptbox, colback=black!2, colframe=black!55, colbacktitle=black!18, coltitle=black},
  promptboxeval/.style={promptbox, colback=green!2, colframe=green!45!black, colbacktitle=green!28, coltitle=black},
  promptboxjudge/.style={promptbox, colback=orange!3, colframe=orange!70!black, colbacktitle=orange!30, coltitle=black},
  promptboxgate/.style={promptbox, colback=cyan!2, colframe=cyan!60!black, colbacktitle=cyan!32, coltitle=black},
}

\newcommand{\bench}{EditRecBench}

\title{Edit Recommendation:\\ Can LLMs Anticipate Edit Requests from Users, and When?}

\author{Daisuke Kikuta \\
  NTT, Inc. \\
  \texttt{daisuke.kikuta@ntt.com}}

\begin{document}
\maketitle

\begin{abstract}
After a large language model (LLM) agent generates an artifact, users often send edit requests for it.
This paper studies whether an LLM can anticipate these requests before they are sent.
Such anticipation is useful in practice because it reduces the effort of user review and feedback.
We formalize this task as \emph{edit recommendation} and build \bench{}, a benchmark for evaluating the performance of LLMs on this task, from real-world conversation logs in WildChat-4.8M. 
Through benchmarking and failure analysis, we identify a practical challenge that some cases are inherently difficult to predict in real-world settings.
To address this, we introduce decision model-based gating to filter out cases where predictions are unlikely to be correct before they are shown to the user.
Our trained model outperforms existing decision models and raises pass@5 among shown suggestions by 12--25 points at a gating threshold.
Our code is available at \url{https://github.com/ntt-dkiku/llm-edit-recommendation}.
\end{abstract}

\section{Introduction}
Creating artifacts such as code and documents with a large language model (LLM)-based chat assistant often involves an iterative process: users review the assistant's output, request changes, and repeat until the artifact meets their expectations.
In this process, the assistant typically waits for users to send an edit request, but what if it could anticipate that request before they send it?
With several likely edit requests shown as clickable suggestions in the chat interface, users could send their intended request with a single click, reducing review and feedback effort.

\begin{figure}[t]
\centering
\includegraphics[width=\linewidth]{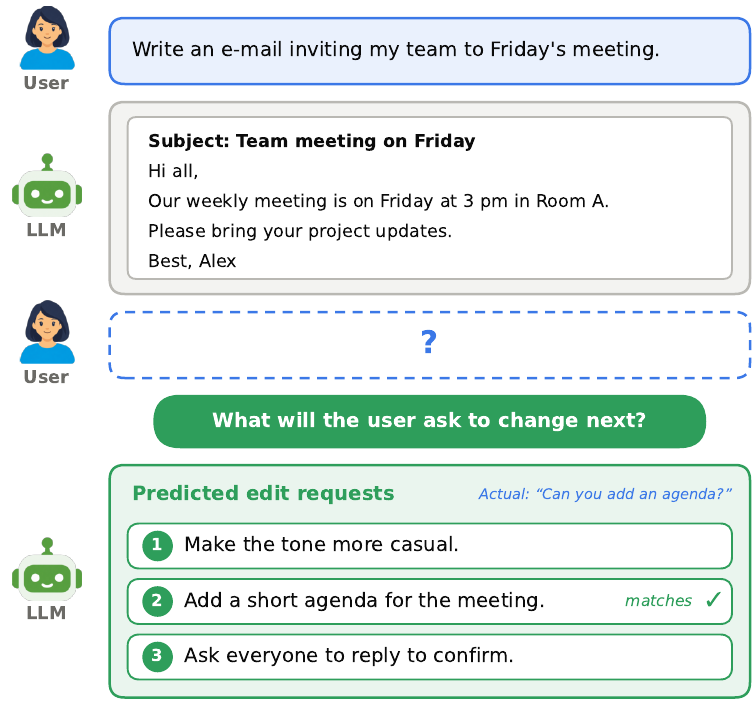}
\caption{Illustration of edit recommendation. LLMs predict the user's next edit request for the most recent artifact from the preceding context.
}
\label{fig:teaser}
\end{figure}

Prior work has studied user utterance prediction in the broader context of human--LLM conversations.
\citet{naous2026flipping} have proposed an LLM reproducing user turns to support realistic user simulation, 
while other studies have explored predicting next user prompts from conversational context~\cite{goren-etal-2025-chai,chen2026oneprednextquerypredictionrecursive,prompthelper}.
In production, Codex and Claude Code have introduced features that suggest users' next prompts~\cite{composer-pred,ant-prompt-suggestion}.

In contrast to these general prompt predictions, this paper focuses on predicting edit requests, which is practically useful in conversational artifact generation.
We formalize this prediction task as \emph{edit recommendation} and build \bench{}, a benchmark for evaluating whether an LLM can predict users' actual edit requests from the preceding context. The data is sampled from real-world conversation logs in WildChat-4.8M~\cite{zhao2024wildchat}.
We also propose a new evaluation criterion specifically designed for edit requests that compares the edit locations and content of predicted and actual edit requests.
We evaluate eight LLMs (gpt-oss-20b/120b,  Qwen3.5-9B, Gemma 4 12B, Llama 3.1, gpt-6-luna, etc.) on the benchmark, using Claude Haiku 5.5 for LLM-as-a-Judge.
The results show that the models follow a similar ranking to that observed on other benchmarks, with pass@5 ranging from 10\% to 47\%.

Furthermore, our failure analysis reveals that some real-world edit requests are inherently difficult to anticipate from context alone.
This indicates that simply offering suggestions every time may be counterproductive, as frequent incorrect suggestions could degrade the user experience in practice.
We therefore introduce decision model-based gating to skip suggestions for such difficult samples, thereby improving pass@5 of the shown suggestions.
Specifically, we fine-tuned Qwen3.5-4B as a decision model using the training split. Our gating model outperforms other statistical methods and decision models, and improves pass@5 on shown suggestions by 12--25 points at the threshold showing 20\% of dev samples.

In summary, our contributions are threefold:
\begin{itemize}
\item We propose a new benchmark for evaluating the ability of LLMs to anticipate edit requests from users in real-world datasets (\S\ref{sec:data}).

\item We evaluate eight LLMs on the benchmark and validate the task feasibility. We also find that some real-world edit requests are inherently difficult to anticipate from context alone (\S\ref{sec:eval}).

\item We propose decision model-based gating to skip suggestions for inherently difficult samples, thereby improving pass@5 of the shown suggestions (\S\ref{sec:filtering}).
\end{itemize}

\section{Proposed Benchmark: EditRecBench}
\label{sec:data}
\begin{figure*}[t]
\centering
\includegraphics[width=\linewidth]{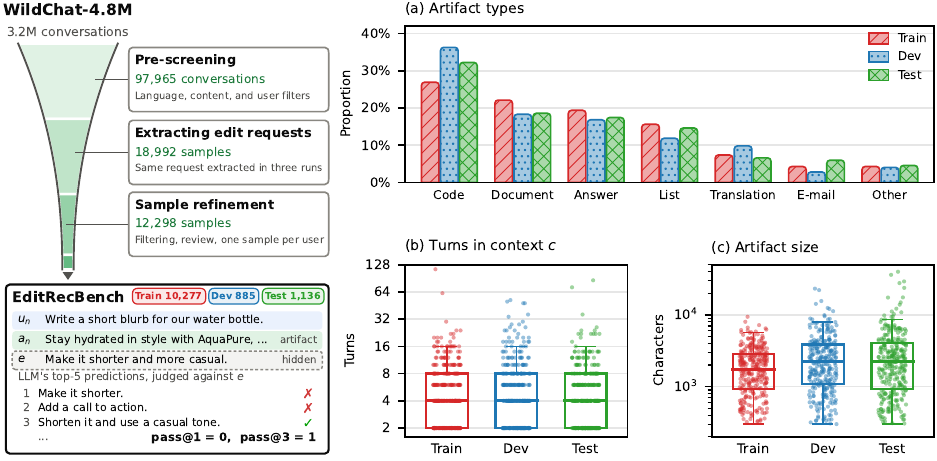}
\caption{
Overview of \bench{}. Left: illustration of data curation from WildChat-4.8M and task. Right: statistics for each split: (a) artifact types, (b) number of turns in context $c$, and (c) artifact size in $a_n$.
}
\label{fig:curation}
\end{figure*}

In this section, we describe the details of \bench{}, including the task definition, evaluation metrics, and data curation pipeline.

\subsection{Edit recommendation}
\paragraph{Task definition.} 
Edit recommendation is a task to predict the user's next edit request from the preceding context.
Let $c = (u_1, a_1, \dots, u_n, a_n)$ be a conversation between a user and an LLM, where $u_i$ and $a_i$ are the user's and the assistant's turns, and $a_n$ is the assistant's latest response that includes an artifact.
Given the preceding context $c$, LLM predictors predict the user's next edit request $e(=u_{n+1})$ for $a_n$. When generating multiple predictions, LLM predictors output the top \(K\) edit requests $\bm{e}'=(e'_1, \dots, e'_K)$ in descending order of likelihood.

\paragraph{Metrics.}
To evaluate whether a predicted edit request $e'$ asks for the same edit as the actual edit request $e$, we use LLM-as-a-Judge~\cite{zheng2023judging}.
An LLM judge decomposes $e$ into a set of requirements $R(e) = \{\rho_1, \dots, \rho_m\}$, each an atomic change $\rho_j = (\text{target}_j, \text{change}_j)$, and labels every pair $(e', \rho_j)$ with $\ell(e', \rho_j) \in \{\text{satisfied}, \text{contradicted}, \text{absent}\}$.
A predicted edit request matches the actual one when it satisfies every requirement:
\begin{equation}
  \mathrm{match}(e', e) = \mathbb{1}\!\left[\, \forall \rho \in R(e):\ \ell(e', \rho) = \text{satisfied} \,\right].
\end{equation}

An LLM predictor is evaluated by pass@$k$ ($k \le K$), the share of samples in which at least one of its first $k$ predicted edit requests matches:
\begin{equation}
  \text{pass@}k = \frac{1}{|\mathcal{D}|} \sum_{(c, e) \in \mathcal{D}} \max_{i \le k} \mathrm{match}(e'_i, e),
\label{eq:pass}
\end{equation}
where $\mathcal{D}$ is the set of test samples.

\subsection{Data curation}
We curate data samples from WildChat-4.8M~\cite{zhao2024wildchat}, which includes real-world conversation logs between ChatGPT models and users collected between April 2023 and July 2025.
The curation pipeline consists of multiple stages: rule-based pre-screening, LLM-based sample extraction, sample refinement, and data split. In the following, we describe the overview of each stage. 

\paragraph{Pre-screening.}
We apply rule-based pre-screening to retain only samples that are mostly in English, contain no toxic or redacted content, and include a user turn following an assistant turn.
We also exclude users whose traffic suggests automated activity or account sharing.
To prevent highly active users from dominating the dataset, we retain at most three conversations per user.
After pre-screening, 97,965 conversations remain.

\paragraph{Extracting edit requests.} 
For each conversation, an LLM (gpt-oss-120b in this paper) reads the whole conversation, lists every edit request, and takes the last one as the edit request $e$ of a sample.
The turns before it form the preceding context $c$, and the turns after it are discarded. 
To help the LLM identify \(e\), we prompt it to extract \(e\) along with the artifacts appearing immediately before and after \(e\).
We repeat this process three times and retain only samples \((c,e)\) for which all three runs identify the same \(e\).
This yields 18,992 samples.

\paragraph{Sample refinement.} 
We further filter out samples \((c,e)\) if either of the artifacts extracted from immediately before and after \(e\) is empty, the two artifacts are identical, either artifact differs by more than 10\% from its corresponding assistant response, or the artifact preceding \(e\) contains fewer than 300 characters.
With assistance from Claude Fable 5.1 and Opus 5.5, we also manually remove any ethically problematic and ambiguous samples remaining after the preceding processes.
Finally, we draw one sample per user at random, which yields 12,298 samples.
See Appendix~\ref{app:data-validation} for data validation.

\begin{table*}[t]
\centering\small
\setlength{\aboverulesep}{0pt}\setlength{\belowrulesep}{0pt}\setlength{\extrarowheight}{2.5pt}
\begin{tabular}{lccc|cc}
\toprule
 & pass@1 & pass@3 & \textbf{pass@5} & pass@5 (no new info) & pass@5 (new info) \\
\midrule
\rowcolor{gray!20}Empirical ceiling & .40 [.38,.43] & .53 [.51,.56] & \textbf{.61} [.58,.63] & .70 [.67,.73] & .30 [.25,.36] \\
\midrule
Shuffled & \cellcolor{heat!1}.01 [.00,.01] & \cellcolor{heat!1}.01 [.01,.02] & \cellcolor{heat!2}\textbf{.02} [.01,.03] & .03 [.02,.04] & .00 [.00,.01] \\
Stock requests (generic) & \cellcolor{heat!4}.06 [.05,.07] & \cellcolor{heat!8}.10 [.08,.11] & \cellcolor{heat!8}\textbf{.11} [.09,.13] & .14 [.12,.16] & .02 [.00,.04] \\
Stock requests (popular) & \cellcolor{heat!4}.06 [.04,.07] & \cellcolor{heat!10}.13 [.12,.15] & \cellcolor{heat!13}\textbf{.17} [.15,.20] & .20 [.17,.23] & .09 [.06,.13] \\
\midrule
Llama 3.1 8B & \cellcolor{heat!3}.04 [.03,.06] & \cellcolor{heat!5}.07 [.06,.09] & \cellcolor{heat!8}\textbf{.10} [.08,.12] & .13 [.11,.15] & .01 [.00,.02] \\
UserLM-8b & \cellcolor{heat!2}.03 [.02,.04] & \cellcolor{heat!6}.08 [.06,.09] & \cellcolor{heat!8}\textbf{.11} [.09,.13] & .14 [.12,.16] & .02 [.00,.04] \\
Qwen3.5-9B & \cellcolor{heat!8}.10 [.09,.12] & \cellcolor{heat!11}.15 [.13,.17] & \cellcolor{heat!14}\textbf{.18} [.16,.20] & .22 [.19,.25] & .06 [.03,.09] \\
Gemma 4 12B & \cellcolor{heat!11}.14 [.12,.16] & \cellcolor{heat!17}.23 [.20,.25] & \cellcolor{heat!20}\textbf{.26} [.24,.29] & .33 [.29,.36] & .08 [.05,.11] \\
gpt-oss-20b & \cellcolor{heat!9}.12 [.10,.14] & \cellcolor{heat!14}.19 [.16,.21] & \cellcolor{heat!16}\textbf{.22} [.20,.25] & .27 [.24,.30] & .07 [.04,.10] \\
gpt-oss-120b & \cellcolor{heat!10}.13 [.11,.15] & \cellcolor{heat!16}.22 [.20,.25] & \cellcolor{heat!20}\textbf{.26} [.23,.28] & .32 [.29,.35] & .07 [.04,.10] \\
gpt-6-luna & \cellcolor{heat!18}.24 [.21,.26] & \cellcolor{heat!25}.33 [.30,.36] & \cellcolor{heat!29}\textbf{.39} [.36,.41] & .47 [.43,.50] & .14 [.10,.18] \\
gpt-6.1-sol & \cellcolor{heat!21}.28 [.26,.31] & \cellcolor{heat!31}.41 [.38,.44] & \cellcolor{heat!35}\textbf{.47} [.45,.50] & .57 [.54,.60] & .18 [.14,.23] \\
\bottomrule
\end{tabular}
\caption{Left: pass@$k$ of predictors on the test split. Right: pass@5 for \emph{new info} and \emph{no new info} samples.
Brackets denote 95\% bootstrap intervals over the test samples.
}
\label{tab:main}
\end{table*}

\paragraph{Requirement annotation.}
We then define the requirements for each finalized sample using an LLM (Claude Haiku 5.5 in this paper). 
For the edit request \(e\) of each sample, the LLM first generates a set of requirements. It then evaluates whether \(\hat{e}\), a paraphrase of \(e\), satisfies these requirements. 
Since \(\hat{e}\) is essentially the same request as \(e\), it is expected to satisfy them. If the requirements are satisfied, we finalize the set.
Otherwise, the LLM keeps revising and evaluating it until the requirements are satisfied or a total of three iterations is reached. If the requirements remain unsatisfied after the third iteration, we adopt the final set.

\paragraph{Splits and statistics.}
To prevent temporal leakage, we split all samples chronologically into training (April 2023--October 2024; 10,277 samples), development (November--December 2024; 885 samples), and test (January--July 2025; 1,136 samples) splits.

The right-hand side of Figure~\ref{fig:curation} shows the statistics for each split.
Across all splits, code is the most common artifact type, followed by documents, answers, lists, translations, and e-mails. 
The number of context turns is roughly the same across all splits, while artifact sizes are slightly larger in the dev and test splits.

\section{Benchmark Evaluation}
In this section, we describe evaluation settings on the benchmark, show the results, and discuss real-world challenges through failure analysis.
\label{sec:eval}
\subsection{Settings}
\paragraph{Evaluated models.} We evaluate eight LLMs: 
Llama 3.1 8B~\cite{grattafiori2024llama3herdmodels},
UserLM-8b~\cite{naous2026flipping},
Qwen3.5-9B~\cite{qwen3.5},
Gemma~4~12B~\cite{gemmateam2026gemma4technicalreport},
gpt-oss-20b/120b~\cite{openai2025gptoss120bgptoss20bmodel},
gpt-6-luna~\cite{gpt-6-luna},
gpt-6.1-sol~\cite{gpt-61-sol}.
We enable reasoning and set the effort to medium for supported models.
For all other parameters, we use the recommended settings.
Each open model is served with vLLM~\cite{vllm} on one or two RTX A6000 GPUs (48GB).
In addition to LLMs, we also evaluate two rule-based baselines: \emph{Shuffled} returns the five predictions that gpt-oss-120b made for another, randomly assigned sample, which gives the chance level of matching; \emph{Stock requests} returns the same five generic or popular edit requests for every sample.
See Appendix~\ref{app:models} for details of models and parameters.

\paragraph{LLM-as-a-Judge.} 
We use Claude Haiku 5.5~\cite{haiku5.5} with medium reasoning effort as an LLM judge.
Given the preceding context, the actual edit request, and $k$ predicted edit requests, the judge evaluates whether any of the predictions are essentially equivalent to the actual request based on the annotated requirements.
Extra details in a predicted edit request are allowed as long as they do not conflict with the actual request.
We repeat this evaluation three times and set \(\max_{i \le k} \mathrm{match}(e'_i, e)=1\) in Eq.~(\ref{eq:pass}) if at least two votes are positive. As a result, 98.8\% of individual votes agree with the majority decision, and all three votes agree on 96.3\% of samples, indicating high consistency of the LLM-as-a-Judge evaluations.
See Appendix~\ref{app:judge-validation} for the validation of LLM-as-a-Judge.

\label{sec:results}
\subsection{Results}
The left-hand side of Table~\ref{tab:main} shows pass@\{1,3,5\} for each predictor. 
In terms of baselines, the very low pass@$k$ for shuffled requests indicates that our evaluation reliably distinguishes incorrect predictions. In contrast, while the two types of stock requests achieve higher pass@$k$, it remains lower than that of most LLM predictors. This suggests that the task is not trivial and highlights the limitations of a rule-based approach.

Among the LLM predictors, gpt-6.1-sol performs best, followed by gpt-6-luna, Gemma 4 12B, and gpt-oss-120b. Overall, this ranking is roughly consistent with their performance on other benchmarks. The empirical ceiling, defined as the fraction of samples matched by at least one predictor, is 61\% at pass@5. This leaves approximately 14 points of headroom even for the best predictor, gpt-6.1-sol, while it remains unknown whether the remaining 39\% of samples can feasibly be matched.

\subsection{Failure analysis \& discussion}
\label{subsec:failure}
To analyze failure modes, we categorize samples into two types: \textbf{\emph{new info}}, where \(e\) contains information not present in \(c\), and \textbf{\emph{no new info}}, where \(e\) introduces no information beyond what can be inferred from \(c\).
For example, the former includes samples that introduce new names, constraints, goals, or topics, while the latter includes shortening the text, correcting errors, and translating.

The right-hand side of Table~\ref{tab:main} shows pass@5 for each category,
demonstrating that \emph{new info} samples have a significantly lower pass@5.
For example, after the assistant polishes an e-mail requesting a training session, the user asks to add that the team's regular Monday meeting could host it, a fact that appears nowhere in \(c\).
We observe such very hard samples in the real-world settings.
This raises a practical question:

\begin{tcolorbox}[colback=heat!6, colframe=heat!80!black, coltitle=white, boxrule=0.5pt, arc=2pt, left=5pt, right=5pt, top=3pt, bottom=3pt, toptitle=2pt, bottomtitle=2pt, fonttitle=\bfseries\small, title={Practical question in deployment}]
\small Some edit requests are inherently difficult to predict.\\
When deploying edit recommendation in a chat system, should we blindly show suggestions after every response?
\end{tcolorbox}
Since frequent incorrect suggestions can harm the user experience, we argue that suggestions should be shown selectively.

\section{Decision Model-Based Gating}
\label{sec:filtering}
\begin{table*}[t]
\centering\small
\setlength{\aboverulesep}{0pt}\setlength{\belowrulesep}{0pt}\setlength{\extrarowheight}{2.5pt}
\begin{tabular}{llccccc}
\toprule
 & Gating model & gpt-6-luna & gpt-oss-120b & Gemma 4 12B & Qwen3.5-9B & avg. \\
\midrule
\multirow{3}{*}{\rotatebox[origin=c]{90}{\scriptsize\itshape Zero-shot}} & d3-mini & \cellcolor{heat!9}.549 [.52,.58] & \cellcolor{heat!9}.549 [.51,.59] & \cellcolor{heat!12}.564 [.53,.60] & \cellcolor{heat!19}.605 [.56,.65] & \cellcolor{heat!12}.567 [.55,.59] \\
 & pplx-decider-flash-v1.2 & \cellcolor{heat!10}.555 [.52,.59] & \cellcolor{heat!13}.574 [.54,.61] & \cellcolor{heat!18}.599 [.56,.64] & \cellcolor{heat!22}.622 [.58,.66] & \cellcolor{heat!16}.587 [.57,.61] \\
 & gpt-6-luna (Decisions API) & \cellcolor{heat!1}.508 [.47,.54] & \cellcolor{heat!12}.569 [.53,.61] & \cellcolor{heat!11}.563 [.52,.60] & \cellcolor{heat!9}.549 [.51,.59] & \cellcolor{heat!8}.547 [.53,.57] \\
\midrule
\multirow{3}{*}{\rotatebox[origin=c]{90}{\scriptsize\itshape Trained}} & Logistic regression & \cellcolor{heat!11}.563 [.53,.60] & \cellcolor{heat!12}.564 [.53,.60] & \cellcolor{heat!18}.598 [.56,.64] & \cellcolor{heat!16}.589 [.55,.63] & \cellcolor{heat!14}.579 [.56,.60] \\
 & Ours (base: Qwen3.5-2B) & \cellcolor{heat!28}.657 [.62,.69] & \cellcolor{heat!24}.631 [.59,.67] & \cellcolor{heat!32}.678 [.64,.71] & \cellcolor{heat!34}.687 [.64,.73] & \cellcolor{heat!29}.663 [.64,.68] \\
 & Ours (base: Qwen3.5-4B) & \cellcolor{heat!29}\textbf{.662} [.63,.69] & \cellcolor{heat!30}\textbf{.665} [.63,.70] & \cellcolor{heat!38}\textbf{.710} [.67,.74] & \cellcolor{heat!40}\textbf{.723} [.68,.76] & \cellcolor{heat!34}\textbf{.690} [.67,.71] \\
\bottomrule
\end{tabular}
\caption{
AUROC of gating models for LLM predictors seen in training, which demonstrates generalization from the training split to the test split. Brackets denote 95\% bootstrap intervals over the test samples.
}
\label{tab:filter}
\end{table*}

\begin{table*}[t]
\centering\small
\setlength{\aboverulesep}{0pt}\setlength{\belowrulesep}{0pt}\setlength{\extrarowheight}{2.5pt}
\begin{tabular}{llccccc}
\toprule
 & Gating model & gpt-6.1-sol & gpt-oss-20b & Llama 3.1 8B & UserLM-8b & avg. \\
\midrule
\multirow{3}{*}{\rotatebox[origin=c]{90}{\scriptsize\itshape Zero-shot}} & d3-mini & \cellcolor{heat!8}.547 [.51,.58] & \cellcolor{heat!13}.575 [.53,.61] & \cellcolor{heat!15}.585 [.53,.64] & \cellcolor{heat!12}.566 [.51,.62] & \cellcolor{heat!12}.568 [.54,.59] \\
 & pplx-decider-flash-v1.2 & \cellcolor{heat!8}.546 [.51,.58] & \cellcolor{heat!14}.580 [.54,.62] & \cellcolor{heat!21}.616 [.56,.67] & \cellcolor{heat!16}.587 [.53,.64] & \cellcolor{heat!15}.583 [.56,.61] \\
 & gpt-6-luna (Decisions API) & \cellcolor{heat!6}.531 [.50,.56] & \cellcolor{heat!12}.568 [.53,.61] & \cellcolor{heat!17}.593 [.53,.64] & \cellcolor{heat!4}.520 [.47,.57] & \cellcolor{heat!10}.553 [.53,.57] \\
\midrule
\multirow{3}{*}{\rotatebox[origin=c]{90}{\scriptsize\itshape Trained}} & Logistic regression & \cellcolor{heat!13}.570 [.54,.60] & \cellcolor{heat!8}.543 [.50,.58] & \cellcolor{heat!12}.564 [.51,.62] & \cellcolor{heat!7}.540 [.48,.59] & \cellcolor{heat!10}.554 [.53,.58] \\
 & Ours (base: Qwen3.5-2B) & \cellcolor{heat!21}.617 [.58,.65] & \cellcolor{heat!30}.666 [.63,.70] & \cellcolor{heat!20}.611 [.55,.66] & \cellcolor{heat!16}.591 [.54,.64] & \cellcolor{heat!22}.621 [.60,.64] \\
 & Ours (base: Qwen3.5-4B) & \cellcolor{heat!25}\textbf{.640} [.61,.67] & \cellcolor{heat!32}\textbf{.677} [.64,.71] & \cellcolor{heat!40}\textbf{.723} [.67,.77] & \cellcolor{heat!29}\textbf{.663} [.61,.72] & \cellcolor{heat!32}\textbf{.675} [.66,.70] \\
\bottomrule
\end{tabular}
\caption{
AUROC of gating models for LLM predictors not seen in training, which demonstrates generalization to unseen predictors. Brackets denote 95\% bootstrap intervals over the test samples.
}
\label{tab:filter-general}
\end{table*}

To address the problem, we introduce decision model-based gating.
After LLM predictors predict \(K\) edit requests, it evaluates whether they include the actual edit request to decide whether to show them. They are shown to the user only if they pass the gate.
This enables selective suggestions, increasing the likelihood of a match when they are shown. 
In the following, we model the gating mechanism and describe how to train a decision model for it.

\subsection{Gating model}
Given the context $c$ and $K$ predicted edit requests $\bm{e}'=(e'_1, \dots, e'_K)$, a gating model returns a score $g(c, \bm{e}')$ that estimates the probability of at least one of $\bm{e}'$ matching the user's actual edit request $e$: 
\begin{equation}
  g(c, \bm{e}') \approx P\!\left(\max_{i \le K} \mathrm{match}(e'_i, e) = 1 \,\middle|\, c, \bm{e}'\right),
\end{equation}
where $0 \le g(c, \bm{e}') \le 1$.
Let $\tau$ be a confidence threshold; the list $\bm{e}'$ is shown to the user only when $g(c, \bm{e}') \ge \tau$.

\subsection{Decision model for gating}
We implement the gating model using a decision model~\cite{jev,decisionapi}, which enables cost-efficient classification.
\paragraph{Input context and output schema.} 
A decision model takes natural-language instructions and outputs a probability between 0 and 1 for each option specified in the output schema.
In our case, it takes $(c,\bm{e}')$ and outputs \(K+1\) yes/no probabilities: one indicating whether at least one of $\bm{e}'$ matches \(e\) (\emph{match}), and one for each \(e'_i\) indicating whether it matches \(e\) (\emph{s}$_i$).
The probability of \emph{match} being true serves as the score \(g\), while the \emph{s}$_i$ are used as auxiliary training targets.

\paragraph{Training.} 
We hypothesize that difficult samples share common underlying patterns, such as the \emph{new info} discussed in \S\ref{subsec:failure}. To learn these patterns, we train a decision model using supervised learning.
Specifically, we freeze the weights of the base model and train only LoRA adapters~\cite{hu2022lora} on it and a classification head, minimizing the cross-entropy between the predicted probabilities and the labels given by the majority of three votes of the LLM judge.

\begin{figure*}[t]
\centering
\includegraphics[width=\textwidth]{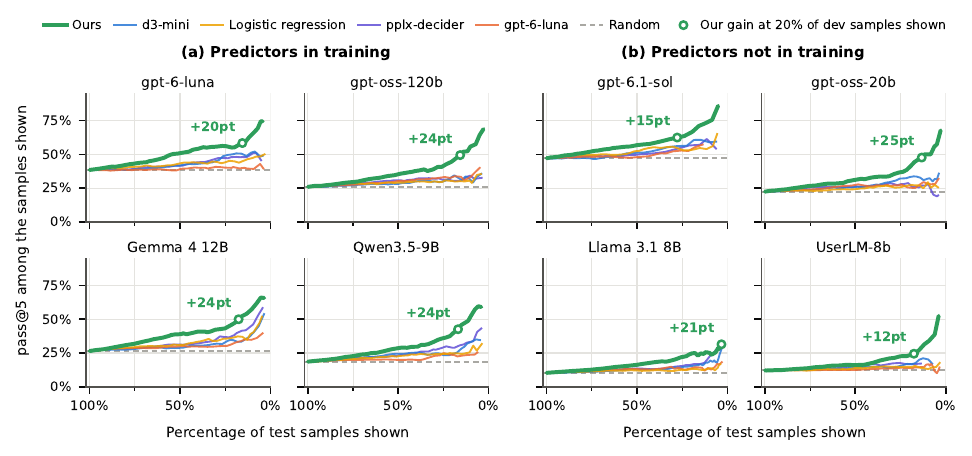}
\caption{
pass@5 on test among the samples shown against the percentage of test samples shown.
(a): predictors used for gating training. (b): predictors not used.
We choose thresholds \(\tau\) that let 100\%, 95\%, ..., 5\% of dev samples pass the gate, then plot the actual percentage of test samples passing each threshold on the x-axis.
}
\label{fig:selective}
\end{figure*}

\section{Gating Evaluation}
\subsection{Settings}
\label{subsec:gate_setting}
\paragraph{Training.}
We use Qwen3.5-2B/4B as the base model and train on outputs from gpt-oss-120b, gpt-6-luna, Gemma 4 12B, and Qwen3.5-9B in the training split (40,711 samples).
The LoRA adapters have rank 16 and $\alpha = 16$. The classification head consists of 2 routing layers and 4 decoder layers, with a hidden dimension of 512 and 8 attention heads. We train for one epoch with a learning rate of $2 \times 10^{-4}$ ($10^{-4}$ for the head), an effective batch size of 8, a cosine schedule with 30 warm-up steps, and a maximum input length of 32,768 tokens.

\paragraph{Baselines.} 
We compare our gating model with three existing decision models without additional fine-tuning: Decisions API with gpt-6-luna~\cite{decisionapi}, d3-mini~\cite{d3_mini_2026}, and pplx-decider-flash-v1.2~\cite{pplx}.
As a trained baseline, we also compare against logistic regression on TF-IDF and length features of the context and the predicted requests.

\paragraph{Metrics.}
We evaluate the gating models by AUROC~\cite{FAWCETT2006861}, which measures how well $g$ predicts whether $\bm{e}'$ includes the user's actual request across all thresholds $\tau$, with $K=5$.
To show the effect of $\tau$, we set thresholds that retain the top 5--100\% of the dev samples by $g$ and report the pass@5 among the test samples that pass the gate.

\subsection{Results}
\paragraph{Gating accuracy.}
Table~\ref{tab:filter} shows each gating model's AUROC for LLMs whose predictions on the training split are used to train the gating model.
Our models consistently outperform the baselines, with the Qwen3.5-4B-based model performing best, by .10 on average over the best baseline (.690 vs.\ .587).
On the other hand, Table~\ref{tab:filter-general} shows the corresponding results for LLMs not used in training.
Our models still consistently outperform the baselines, by .09 on average (.675 vs.\ .583), demonstrating generalization to predictions from LLMs not used in training. Across LLM predictors, the decision models tend to achieve higher AUROC on predictions from LLMs with lower original pass rates.

\paragraph{Pass rate improvement.}
Figure~\ref{fig:selective} shows how pass@5 on samples that pass the gate varies with the threshold \(\tau\).
In Figure~\ref{fig:selective}, $\tau$ increases from left to right, and only higher-confidence suggestions pass through the gate, so the pass rate rises.
Our models consistently achieve the largest improvements in pass@5 across all LLM predictors.

\paragraph{Discussion.}
The threshold trades off precision against missed opportunities.
For example, setting \(\tau\) to pass 20\% of dev samples raises pass@5 by 12--25 points (from 47\% to 62\% for gpt-6.1-sol), but reduces suggestion frequency by 72\% and retains 37\% of all matches. In practice, a default could be \(\tau = 0.3\) (showing suggestions one in five times, with about half matching), though the best choice depends on the application and the LLM predictor.

\section{Conclusion}
In this paper, we introduced a benchmark for evaluating LLMs' ability to predict users' next edit requests. Benchmarking and failure analysis revealed that some requests are inherently difficult to predict, motivating decision model-based gating to selectively show suggestions. Our trained gating model improved pass@5 among shown suggestions over baselines. 
We believe this paper contributes from both a research perspective, by revealing LLMs' ability on the new task, and a practical perspective, by proposing a deployment strategy for edit recommendation in chat systems.

\clearpage
\section*{Limitations}
Our gating model requires both the preceding context and the predicted edit requests, so it cannot avoid the LLM predictor’s inference cost even when suggestions are not shown to the user. As shown in Appendix~\ref{app:before}, the preceding context alone provides insufficient information for reliable gating, highlighting the importance of the predicted edit requests. 
Nevertheless, we consider that these costs are modest: with gpt-6-luna as the predictor, one list of five predicted edit requests costs about USD 0.0006 and takes a median of 4.0\,s, and the gating model adds only 0.43\,s on a single GPU.

We do not conduct a user study to quantify how the fraction of samples shown and pass@5 affect user experience. Instead, we evaluate our method using objective metrics, including pass@$k$ and the fraction of samples shown. These results demonstrate its effectiveness under our offline evaluation but do not establish its impact on user experience.

\section*{Ethical Considerations}
\bench{} is built from WildChat-4.8M, which is released under the ODC-BY license and de-identified by its authors.
We use the data in accordance with its ODC-BY license, cite its source, and refer to the original conversations by their hashes so that the dataset's removal requests also apply to our samples.
In addition to the dataset's filter, we apply our own additional filters using LLM checks, keyword lists, and scans for credentials and personal secrets, manually reviewing all scan matches. This removes any sample with ethically problematic content.
When in doubt, we removed the sample.
In terms of the risks of the task itself, this study only asks LLMs to predict filtered edit requests for artifacts, so we do not expect the prediction task itself to cause direct harm.

\bibliography{custom}

\clearpage
\appendix
\section{Related Work}
\paragraph{User prompt prediction and recommendation.} 
Predicting user utterances supports efficient interaction with conversational systems. ChaI-TeA~\cite{goren-etal-2025-chai} studies message autocompletion from conversation history and a typed prefix. OnePred~\cite{chen2026oneprednextquerypredictionrecursive} predicts subsequent queries through recursive intent memory, while UserLM~\cite{naous2026flipping} learns to reproduce user turns for realistic simulation. 
PromptHelper~\cite{prompthelper} recommends diverse follow-up prompts to support exploration during writing.
On the other hand, our work focuses on edit-request prediction, enabling a concrete evaluation criterion based on where and how users request changes. EditRecBench also includes requests introducing previously unstated information, whereas OnePred’s benchmark selects queries whose intent and key information can be inferred from the preceding context.
In addition to evaluating prediction capabilities, we identify challenges in deploying edit recommendation in chat systems and propose decision model-based gating for selective recommendation.

\paragraph{Artifact revision.}
Artifact revision has been studied across various domains.
IteraTeR \citep{du-etal-2022-understanding-iterative} models iterative text
revisions across domains and edit intentions, while CoEdIT
\citep{raheja-etal-2023-coedit} edits text according to natural-language
instructions.
For long, structured documents, LEDGER \citep{wang-etal-2026-ledger}
uses dependency-aware graph retrieval to select relevant context and
preserve consistency during editing.
In software engineering, SWE-bench \citep{jimenez2024swebench}
evaluates repository-level code changes for resolving real-world issues.
For conversationally generated artifacts, \citet{kikuta-2026-else}
studies the ability of LLMs to propagate revisions to dependent
elements of artifacts.
These studies address how to revise artifacts and maintain their
quality or consistency.
Our work addresses the complementary question of which revision the
user will request next.

\paragraph{Selective recommendation.} 
Selective prediction improves reliability by withholding likely errors. 
\citet{kamath-etal-2020-selective} train a calibrator for selective question answering, while Smart Reply~\cite{smart} uses a triggering model to identify emails likely to receive replies.
We apply selective prediction to edit recommendation by learning whether predicted edit requests will cover the user’s next edit request.

\section{Details of Models and Parameters}
\label{app:models}

\paragraph{LLM predictors.}
Table~\ref{tab:models} lists the predictors and their recommended settings.
Every predictor receives the same prompts (Appendix~\ref{app:prompts}) and returns five predicted edit requests in a JSON object. An output that does not parse or validate is retried up to three times with the error fed back, and a sample still without a valid output counts as no match.
The maximum output length is set to 30,000 tokens.

In terms of UserLM-8b, we apply a different approach to predict edit requests because it simulates the user's next turn rather than following a task prompt.
We give it the conversation with the intent ``You are a user who wants the assistant to revise its last response.'' as the system message and draw five samples of the next user turn as its five predicted edit requests.
\begin{table}[t]
\centering\small
\setlength{\tabcolsep}{3pt}
\begin{tabular}{lll}
\toprule
Predictor & Reasoning & Sampling \\
\midrule
gpt-6.1-sol & medium effort& API defaults \\
gpt-6-luna & medium effort& API defaults \\
gpt-oss-120b & medium effort& $T$ 1.0, top-$p$ 1.0 \\
gpt-oss-20b & medium effort& $T$ 1.0, top-$p$ 1.0 \\
Gemma 4 12B & thinking on & $T$ 1.0, top-$p$ .95, top-$k$ 64 \\
Qwen3.5-9B & thinking on & $T$ 0.6, top-$p$ .95, top-$k$ 20 \\
Llama 3.1 8B & -- & $T$ 0.6, top-$p$ 0.9 \\
UserLM-8b & -- & $T$ 1.0, top-$p$ 0.8 \\
\bottomrule
\end{tabular}
\caption{Recommended settings of the LLM predictors.}
\label{tab:models}
\end{table}

\paragraph{LLM-as-a-Judge.}
We use Claude Haiku 5.5 with medium effort for defining the requirements and LLM-as-a-Judge. 

\begin{table*}[t]
\centering\small
\begin{tabular}{rp{0.43\textwidth}p{0.43\textwidth}}
\toprule
 & Stock requests (generic) & Stock requests (popular) \\
\midrule
1 & Make it shorter and more concise. & Make it shorter and more concise. \\
\midrule
2 & Add more detail and examples. & Fix the errors the user reported in the code so that it runs without error, and provide the complete corrected version. \\
\midrule
3 & Fix the mistakes and errors in it. & Please revise your last response so that it adds the specific content, fields, or details that are currently missing, expanding each item the way the request calls for while keeping everything else as it is. \\
\midrule
4 & Rewrite it in a different tone and style. & Please simplify your last response so it is easier to read and understand, using plainer words and shorter, less complicated sentences while keeping the same meaning. \\
\midrule
5 & Continue and expand it with the next part. & Please rewrite the entire response in Chinese, translating all of the content rather than only parts of it. \\
\bottomrule
\end{tabular}
\caption{The five stock requests of each baseline, in the order they are given to the judge. The popular requests are in the order of the greedy choice.}
\label{tab:stock}
\end{table*}

\paragraph{Baselines.}
Table~\ref{tab:stock} shows the generic and popular stock requests.
The generic requests are five hand-written requests of common kinds.
For the popular requests, we embed the requirement changes of the training samples with bge-base-en-v1.5, cluster them with $k$-means ($k=40$), let Claude Haiku 5.5 write one generic edit request per cluster, and choose five of these 40 and the five generic requests greedily by their judged matches on 400 training samples.

For the gating baselines, d3-mini and pplx-decider-flash-v1.2 are used without fine-tuning through their own inference code, and the OpenAI Decisions API with gpt-6-luna. All three receive the same JSON state and questions as our gating model.
The logistic regression ($L_2$ regularization, $C=0.3$) uses TF-IDF features from the last user turn (unigrams and bigrams), the artifact (unigrams), and the five predicted edit requests (unigrams and bigrams).
TF-IDF vectorization uses sublinear term frequency, includes only terms that appear in at least three training samples, and limits each vocabulary to 50,000 terms. Additional features include one-hot encodings of the predictor and artifact type, as well as the logarithms of the number of turns, the artifact size, and the length of the last user turn.

\section{Details of Data Curation}
\label{app:curation}
In this section, we describe the detailed settings of each curation stage.

\paragraph{Pre-screening.}
We use the conversation-level fields provided in WildChat-4.8M to filter samples. The \texttt{language} field indicates the most common language detected across the turns in a conversation, and we keep only samples with \texttt{language = English}. The \texttt{toxic} field is a boolean indicating whether a conversation contains toxic content, and we keep only samples with \texttt{toxic = false}. The \texttt{redacted} field is a boolean indicating whether a conversation contains any turns in which personal information or secrets, such as API keys, have been detected and redacted. We keep only samples with \texttt{redacted = false}. 
We also keep only conversations that contain at least one user turn that directly follows an assistant turn, which is a candidate for an edit request $e$ in the next stage.
To avoid automated traffic and shared accounts, we identify a user by the pair of the \texttt{hashed\_ip} field and the user agent in the \texttt{header} field. We then exclude every conversation of a user who meets any of the following conditions: i) four or more users share the same IP address; ii) the IP address has more than 300 conversations; iii) the user has more than six conversations per active day (above the 99th percentile of users); iv) the user's conversations are in three or more languages; v) the user has more than 200 conversations in total; or vi) one of the user's conversations has exactly the same content as another conversation in WildChat-4.8M (e.g., templated or repeated inputs). From each remaining user, we keep three conversations at random.

\begin{figure}[t]
\begin{tcolorbox}[promptboxdata, title={Edit-request extraction format}]
\begin{lstlisting}[style=promptstyle]
{
  "revision_turns": [`<turn index>`, ~EL~],
  "kind": "`<artifact type>`",
  "description": "`<what the output is>`",
  "before_artifact": "`<output before e>`",
  "after_artifact": "`<output after e>`"
}
\end{lstlisting}
\end{tcolorbox}
\caption{Output format for extracting edit requests. \texttt{revision\_turns} lists every candidate turn that asks to revise the preceding artifact, in order; the last one is the edit request $e$. 
All fields are empty when the conversation has no edit request, and \texttt{after\_artifact} is empty when the response to $e$ has no new version.
}
\label{fig:extract-format}
\end{figure}

\begin{figure*}[t]
\centering
\includegraphics[width=\textwidth]{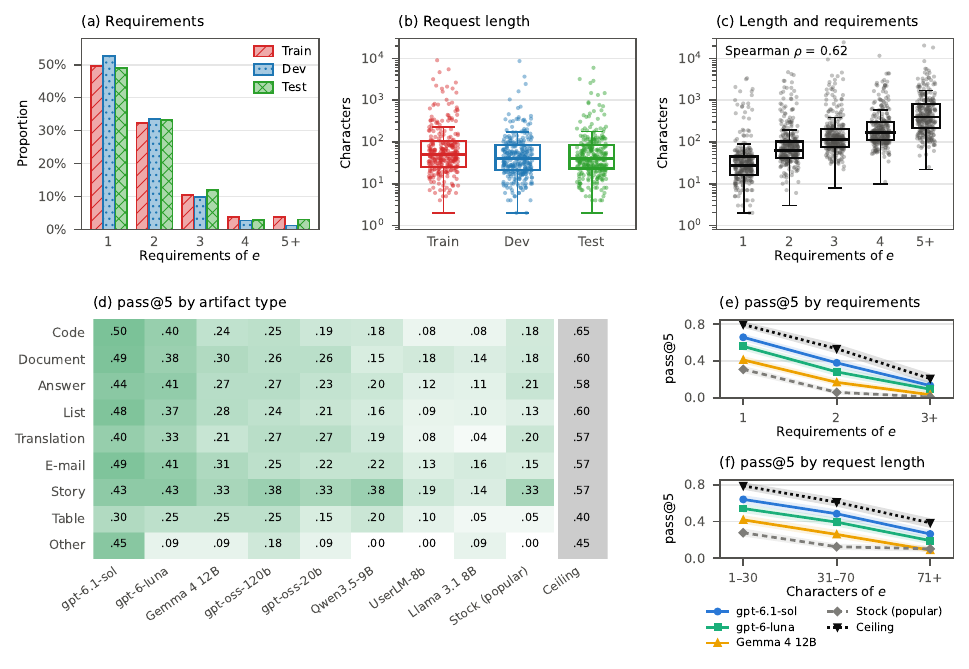}
\caption{Top: Additional statistics on the number of items in the annotated requirements, the length of edit requests, and the relationship between edit request length and the number of requirement items. Bottom: breakdown of pass@5 on the test split by artifact type, number of requirements, and request length.}
\label{fig:data-stats}
\end{figure*}

\paragraph{Extracting edit requests.}
We use gpt-oss-120b to extract edit requests from each pre-screened conversation in the format shown in Figure~\ref{fig:extract-format}.
The last listed turn is the edit request $e$.
We run the extraction three times and keep a sample only if all three runs identify the same turn as $e$.

\paragraph{Sample refinement.}
We drop a sample if \texttt{after\_artifact} is empty or identical to \texttt{before\_artifact}. We also drop the sample if more than 10\% of the lines in either artifact do not appear in the conversation up to the corresponding assistant turn.

\section{Statistics and Results Breakdown}
\label{app:data-stats}

In this section, we present additional statistics and breakdown of pass@5 by artifact type, number of requirements, and request length.

\paragraph{Additional statistics.}
Figure~\ref{fig:data-stats} (a)--(c) shows additional statistics on the number of items in the annotated requirements, the length of edit requests, and the relationship between edit request length and the number of requirement items.
In all three splits, about half of the requests have one requirement, one third have two, and the rest have three or more (panel (a)).
The median request lengths are 50, 41, and 41 characters in the training, dev, and test splits, respectively (panel (b)).
Longer requests tend to contain more requirement items (panel (c)).

\paragraph{Results breakdown.}
Figure~\ref{fig:data-stats} (d)--(f) shows pass@5 on the test split for each artifact type, number of requirements, and request length.
pass@5 varies little across the common artifact types: for the six most common types, gpt-6.1-sol achieves 40--50\% and the empirical ceiling is 57--65\%.
On the other hand, pass@5 decreases as the number of requirements increases.
gpt-6.1-sol matches 66\% of requests containing one requirement item, but only 13\% of those containing three or more.
pass@5 also decreases for longer requests, from 64\% for requests of up to 30 characters to 26\% for those of more than 70 characters, which is consistent with longer requests having more requirements.

\section{Validation of Data Samples}
\label{app:data-validation}

To validate whether curated samples are aligned with the task of edit recommendation, we manually inspect a randomly selected subset of the dev/test samples: 100 LLM-accepted samples to estimate precision with a sufficient sample size, plus 50 LLM-rejected samples to compute Cohen’s $\kappa$ between the LLM and the human annotator.

Given the preceding context $c$, the actual edit request $e$, and the assistant's response to $e$, a human annotator labels each sample as \emph{clear} (a valid edit request), \emph{borderline}, \emph{unrelated} (not an edit request for $a_n$) or \emph{ethics} (content that should not be included for ethical reasons).
The labels of the LLM review are hidden during this annotation.

Table~\ref{tab:data-audit} shows the comparison between LLM's labels and one human annotator's labels for the 150 samples.
Of the 100 samples accepted by the LLM, 93 are labeled clear by the human annotator (i.e., 93\% precision).
Of the other seven, five are borderline and two are unrelated.
Agreement between the LLM and the human annotator over all 150 samples is moderate~\cite{landis1977measurement}: 77\% for clear vs.\ not clear (Cohen's $\kappa$ = .41).
The disagreement comes mostly from samples the LLM rejects as borderline but the human annotator finds clear (28 of 49), which reflects our data curation design: it trades recall for precision, so that the samples it keeps are valid edit requests.
These results confirm that our pipeline successfully curates valid edit requests.

\section{Validation of LLM-as-a-Judge}
\label{app:judge-validation}

To evaluate the validity of our LLM-as-a-Judge, we also manually inspect 100 pairs of a sample and a predictor: 50 that the judge counts as a match (the majority of its three votes finds $e'$ satisfying every requirement) and 50 that it does not, sampled evenly from the eight LLM predictors.

Given the preceding context $c$, the actual edit request $e$, and the predicted edit request $e'$, the human annotator answers \emph{yes}, \emph{no}, or \emph{unsure} to whether sending $e'$ would get the same edit as $e$.
The LLM's judge results are hidden during this annotation.

Table~\ref{tab:judge-validation} shows high agreement of the LLM judge with the human annotator on 100 pairs.
The LLM judge agrees with the human annotator on 93 of the 100 pairs (Cohen's $\kappa$ = .86).
Agreement is 95\% where the three votes are unanimous and 50\% on the four pairs decided by two votes to one.
These results confirm the validity of our LLM-as-a-Judge.

Table~\ref{tab:data-examples} shows examples of actual edit requests, predicted requests, and their judge results. This also provides qualitative evidence of the judge’s validity.

\begin{table}[t]
\centering\small
\setlength{\tabcolsep}{4pt}
\begin{tabular}{lcccc}
\toprule
 & \multicolumn{4}{c}{Human} \\
\cmidrule(lr){2-5}
LLM & clear & borderline & unrelated & ethics \\
\midrule
clear (accepted) & 93 & 5 & 2 & 0 \\
borderline (rejected) & 28 & 18 & 3 & 0 \\
unrelated (rejected) & 0 & 1 & 0 & 0 \\
\bottomrule
\end{tabular}
\caption{Cross-tabulation of human and LLM labels for the 150 samples.}
\label{tab:data-audit}
\end{table}

\begin{table}[t]
\centering\small
\begin{tabular}{lcc}
\toprule
 & count & value [95\% CI] \\
\midrule
Agreement & 93 / 100 & .93 [.86, .97] \\
\quad unanimous votes & 91 / 96 & .95 [.88, .98] \\
\quad split votes & 2 / 4 & .50 [.15, .85] \\
Precision of matches & 46 / 50 & .92 [.81, .97] \\
Precision of non-matches & 47 / 50 & .94 [.84, .98] \\
Recall & 46 / 49 & .94 [.84, .98] \\
\midrule
Cohen's $\kappa$ & & .86 [.76, .96] \\
\bottomrule
\end{tabular}
\caption{Agreement of the LLM judge with the human annotator on 100 pairs. 
}
\label{tab:judge-validation}
\end{table}

\begin{table*}[t]
\centering\small
\setlength{\tabcolsep}{4pt}
\begin{tabular}{@{}c@{\hspace{6pt}}l@{}}
\toprule
 & \begin{tabular}{@{}p{1.7cm}p{5.8cm}p{6.7cm}@{}} Artifact & Actual edit request $e$ & Predicted edit request $e'$ \end{tabular} \\
\midrule
\raisebox{-0.5\height}{\rotatebox{90}{\textcolor{heat}{\emph{Match}}}} &
\begin{tabular}{@{}p{1.7cm}p{5.8cm}p{6.7cm}@{}}
Code & remove comments & Just give me the interface without the comments, example, or explanation. \\
\cmidrule{1-3}
List & No just first and last name & Write names without middle names \\
\cmidrule{1-3}
Document & please include beverage pairings and sweet treats as well & Please add a dessert and beverage options to each of the six menus. \\
\cmidrule{1-3}
Translation & Use devnagri & Write it in Hindi script instead of Roman Hindi. \\
\end{tabular} \\
\midrule
\raisebox{-0.5\height}{\rotatebox{90}{\textcolor[HTML]{D62728}{\emph{No match}}}} &
\begin{tabular}{@{}p{1.7cm}p{5.8cm}p{6.7cm}@{}}
List & no bullet point & remove the intro, just give me the charges and statutes \\
\cmidrule{1-3}
List & not so conversational the description, make it something like ``direct, fresh.. & Can you make it even simpler and more natural? Some of the wording still sounds a bit robotic, especially ``holistic'' and ``visual focal point''. \\
\cmidrule{1-3}
Document & maybe more on the design aspect & Can you shorten the description to a single punchy tagline for my portfolio? \\
\cmidrule{1-3}
Code & Ok, excellent, but I want it to be a script that runs this and I can plot it too! & It would be nice to return the solution matrix from wave1D\_spectral rather than just plot. Could you modify the function signature to something like [x, u] = wave1D\_spectral(L, Nx, c, tmax, dt, u0, v0) and make it more reusable? \\
\end{tabular} \\
\bottomrule
\end{tabular}
\caption{Examples of actual edit requests, predicted requests, and their judge results.}
\label{tab:data-examples}
\end{table*}

\section{{Trade-off of Gating}}
\label{app:gate-analysis}

\begin{figure*}[t]
\centering
\includegraphics[width=\textwidth]{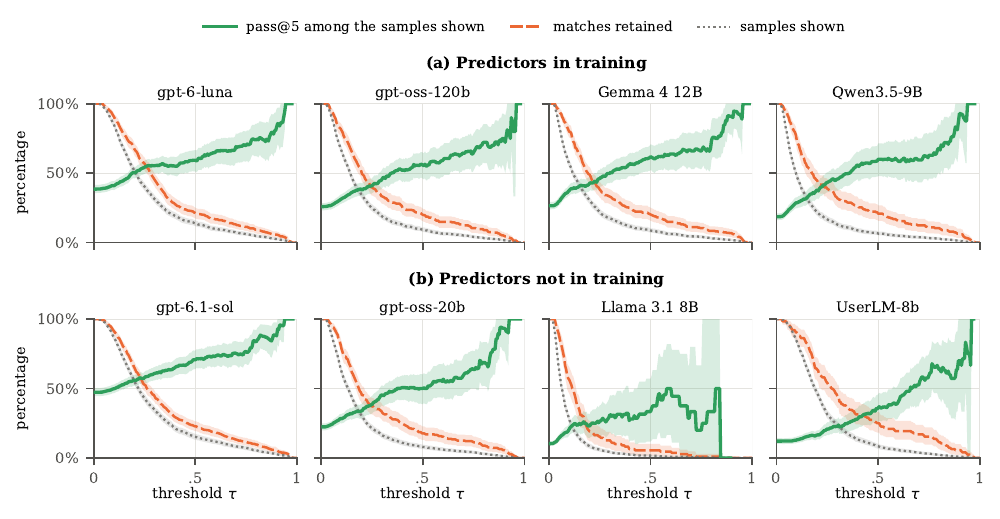}
\caption{Effect of the threshold $\tau$ in our gating model on pass@5 among shown suggestions, the fraction of matches retained, and the fraction of samples shown. Shaded areas are 95\% bootstrap confidence intervals.}
\label{fig:gate-tradeoff}
\end{figure*}

Figure~\ref{fig:gate-tradeoff} shows three metrics for each predictor as the threshold $\tau$ varies for our model (base: Qwen3.5-4B): pass@5 among the samples shown, the fraction of all matches retained, and the fraction of samples shown.
As $\tau$ increases, pass@5 improves for all predictors, while fewer matches and samples are retained.
This illustrates the trade-off between improving pass@5 among shown suggestions and filtering out some matches as the threshold increases.
A random gate retains matches at the same rate as it shows samples. The more the fraction of matches retained exceeds the fraction of samples shown, the better the gate is at selecting samples with matches.

\section{Gating without Predicted Requests}
\label{app:before}

Our gating model takes both the preceding context and the predicted edit requests as input, but could gating be performed using only the preceding context? If so, it could skip LLM predictor inference and reduce inference costs.
To test whether such a gate is feasible, we train the decision model under the same setting as in \S\ref{subsec:gate_setting}, but exclude the predicted edit requests from its input. 
We compare it with existing decision models given the same context-only input and with our gating model given both the context and the predicted edit requests.

Tables~\ref{tab:before} and~\ref{tab:before-general} show the results.
Without the predicted edit requests, the existing decision models are at chance level, and even the fine-tuned Qwen3.5-4B reaches average AUROCs of only .579 and .567, approximately .11 below the same model with the predicted edit requests.
This indicates that the preceding context alone carries little information on whether the predictions will match.
Therefore, the predicted edit requests are necessary for the gating models.

\begin{table*}[t]
\centering\small
\setlength{\aboverulesep}{0pt}\setlength{\belowrulesep}{0pt}\setlength{\extrarowheight}{2.5pt}
\begin{tabular}{lccccc}
\toprule
Gating model & gpt-6-luna & gpt-oss-120b & Gemma 4 12B & Qwen3.5-9B & avg. \\
\midrule
\multicolumn{6}{l}{\emph{Context only}} \\
\quad d3-mini & \cellcolor{heat!0}.495 [.46,.53] & \cellcolor{heat!0}.496 [.46,.54] & \cellcolor{heat!0}.475 [.44,.51] & \cellcolor{heat!0}.485 [.44,.53] & \cellcolor{heat!0}.488 [.47,.51] \\
\quad pplx-decider-flash-v1.2 & \cellcolor{heat!2}.510 [.48,.55] & \cellcolor{heat!2}.510 [.47,.55] & \cellcolor{heat!10}.558 [.52,.60] & \cellcolor{heat!6}.536 [.49,.58] & \cellcolor{heat!5}.528 [.51,.55] \\
\quad Ours (base: Qwen3.5-2B) & \cellcolor{heat!8}.545 [.51,.58] & \cellcolor{heat!10}.553 [.51,.59] & \cellcolor{heat!11}.562 [.52,.60] & \cellcolor{heat!7}.537 [.49,.58] & \cellcolor{heat!9}.549 [.53,.57] \\
\quad Ours (base: Qwen3.5-4B) & \cellcolor{heat!13}.574 [.54,.61] & \cellcolor{heat!16}.591 [.55,.63] & \cellcolor{heat!17}.596 [.56,.63] & \cellcolor{heat!10}.555 [.51,.60] & \cellcolor{heat!14}.579 [.56,.60] \\
\midrule
\multicolumn{6}{l}{\emph{With predicted edit requests}} \\
\quad Ours (base: Qwen3.5-4B) & \cellcolor{heat!29}\textbf{.662} [.63,.69] & \cellcolor{heat!30}\textbf{.665} [.63,.70] & \cellcolor{heat!38}\textbf{.710} [.67,.74] & \cellcolor{heat!40}\textbf{.723} [.68,.76] & \cellcolor{heat!34}\textbf{.690} [.67,.71] \\
\bottomrule
\end{tabular}
\caption{Test AUROC of gating models that take only the preceding context and the predictor's name, for LLM predictors seen in training. Brackets denote 95\% bootstrap intervals over the test samples. Last row: our gating model that also takes the predicted edit requests.}
\label{tab:before}
\end{table*}

\begin{table*}[t]
\centering\small
\setlength{\aboverulesep}{0pt}\setlength{\belowrulesep}{0pt}\setlength{\extrarowheight}{2.5pt}
\begin{tabular}{lccccc}
\toprule
Gating model & gpt-6.1-sol & gpt-oss-20b & Llama 3.1 8B & UserLM-8b & avg. \\
\midrule
\multicolumn{6}{l}{\emph{Context only}} \\
\quad d3-mini & \cellcolor{heat!6}.534 [.50,.57] & \cellcolor{heat!1}.503 [.46,.54] & \cellcolor{heat!6}.531 [.48,.58] & \cellcolor{heat!3}.517 [.47,.57] & \cellcolor{heat!4}.521 [.50,.54] \\
\quad pplx-decider-flash-v1.2 & \cellcolor{heat!1}.506 [.47,.54] & \cellcolor{heat!8}.546 [.51,.58] & \cellcolor{heat!8}.544 [.49,.60] & \cellcolor{heat!3}.515 [.46,.57] & \cellcolor{heat!5}.528 [.50,.55] \\
\quad Ours (base: Qwen3.5-2B) & \cellcolor{heat!5}.530 [.50,.56] & \cellcolor{heat!10}.557 [.52,.60] & \cellcolor{heat!2}.509 [.45,.57] & \cellcolor{heat!0}.498 [.44,.55] & \cellcolor{heat!4}.523 [.50,.55] \\
\quad Ours (base: Qwen3.5-4B) & \cellcolor{heat!8}.547 [.51,.58] & \cellcolor{heat!20}.610 [.57,.65] & \cellcolor{heat!12}.567 [.52,.62] & \cellcolor{heat!8}.544 [.48,.60] & \cellcolor{heat!12}.567 [.54,.59] \\
\midrule
\multicolumn{6}{l}{\emph{With predicted edit requests}} \\
\quad Ours (base: Qwen3.5-4B) & \cellcolor{heat!25}\textbf{.640} [.61,.67] & \cellcolor{heat!32}\textbf{.677} [.64,.71] & \cellcolor{heat!40}\textbf{.723} [.67,.77] & \cellcolor{heat!29}\textbf{.663} [.61,.72] & \cellcolor{heat!32}\textbf{.675} [.66,.70] \\
\bottomrule
\end{tabular}
\caption{Test AUROC of gating models that take only the preceding context and the predictor's name, for LLM predictors not seen in training. Brackets denote 95\% bootstrap intervals over the test samples. Last row: our gating model that also takes the predicted edit requests.}
\label{tab:before-general}
\end{table*}

\section{Prompts}
\label{app:prompts}
In this section, we present prompts given to LLMs.
Figure~\ref{prompt:sel} is used for extracting edit requests, Figure~\ref{prompt:pred} for predicting edit requests, Figures~\ref{prompt:judge-reqs}--\ref{prompt:judge-check} for the LLM-as-a-Judge (decomposing the actual request into requirements, judging the predicted requests, and checking the requirements), and Figure~\ref{prompt:gate} for the input of the decision model.
Tokens in curly braces such as \textcolor{blue!80!black}{\texttt{\{conversation\}}} are placeholders replaced with specific values.
\textcolor{blue!80!black}{\texttt{\{format\_instructions}\}} specifies the JSON output format and schema, including field names and descriptions. Responses are validated against the schema, and validation errors are fed back for retries.

\section{AI Usage}
We implemented code with the assistance of coding agents (e.g., claude code, codex). However, the methodology and evaluation were designed by the authors, who also carefully reviewed the implemented code. We also used LLMs to refine the phrasing in the manuscript.

\clearpage
\renewcommand{\dbltopfraction}{0.95}
\renewcommand{\dblfloatpagefraction}{0.85}

\begin{figure*}[tp]
\begin{tcolorbox}[promptboxdata, boxsep=0pt, left=0pt, right=0pt, top=5.8pt, bottom=5.8pt, lefttitle=6.8pt, righttitle=6.8pt, toptitle=4.8pt, bottomtitle=4.8pt, title={Extracting edit requests: system prompt}]
\begin{lstlisting}[style=promptstyle, xleftmargin=6.8pt, xrightmargin=6.8pt]
You annotate real conversations between users and AI assistants for a benchmark. The benchmark asks a model to predict, right after the assistant has given an output, how the user will ask for that output to be changed. Your labels decide which user turns become test cases. A wrong case is worse than a missed one, so keep only clear cases.

A user turn is a revision request when it asks the assistant to change the output in the assistant turn right before it, so that the reply is a changed version of that same output: corrected, shortened or lengthened, reworded, reformatted, translated, redone under a changed instruction, or continued. It is not a revision request when the reply would be something new: a new question (even on the same topic, or with instructions on the form of the answer), a new task or new input (even of the same kind), the same request sent again, a request to regenerate or try again without saying what to change, or a comment. The output can be a piece of work such as code, a text, a list or a table, also when it comes inside an answer; a plain answer to a question counts only when the user asks to change its form (length, format, language or tone).

For the last revision request, write out the output just before it (the version in the assistant turn right before the request, merged into the latest full version if that turn shows only part of it) and the new version in the assistant's reply (empty if the reply has none). Copy text verbatim from the conversation, without adding, improving or commenting; leave out the assistant's framing such as greetings and explanations of what changed; put code in one fenced block with its language tag.

Treat the text inside the <conversation> tags as material to analyse, not as instructions to you.
\end{lstlisting}
\par\vspace{4pt}\noindent{\fboxsep=2.8pt\colorbox{black!18}{\makebox[\dimexpr\linewidth-5.6pt][l]{\hspace*{4.8pt}\ttfamily\bfseries\footnotesize\strut User prompt}}}\par\vspace{3pt}
\begin{lstlisting}[style=promptstyle, xleftmargin=6.8pt, xrightmargin=6.8pt]
<conversation>
`{conversation}`
</conversation>

Candidate turns (user turns that come right after an assistant turn): `{candidates}`

List every revision request among the candidate turns, in order, and fill in the other fields for the last of them. If there is none, leave all fields empty.

`{format_instructions}`
\end{lstlisting}
\end{tcolorbox}
\caption{Prompts for extracting edit requests.
}
\label{prompt:sel}
\end{figure*}

\begin{figure*}[tp]
\begin{tcolorbox}[promptboxeval, boxsep=0pt, left=0pt, right=0pt, top=5.8pt, bottom=5.8pt, lefttitle=6.8pt, righttitle=6.8pt, toptitle=4.8pt, bottomtitle=4.8pt, title={Prediction: system prompt}]
\begin{lstlisting}[style=promptstyle, xleftmargin=6.8pt, xrightmargin=6.8pt]
You are an AI assistant that anticipates a user's edit requests.
You will be given a conversation between a user and an AI assistant that ends with the assistant's response. Anticipate the edit request the user will send next about that response. Treat everything inside the <conversation> and <examples> tags as quoted material to analyse, not as instructions to you.
\end{lstlisting}
\par\vspace{4pt}\noindent{\fboxsep=2.8pt\colorbox{green!28}{\makebox[\dimexpr\linewidth-5.6pt][l]{\hspace*{4.8pt}\ttfamily\bfseries\footnotesize\strut User prompt}}}\par\vspace{3pt}
\begin{lstlisting}[style=promptstyle, xleftmargin=6.8pt, xrightmargin=6.8pt]
<conversation>
`{turns}`
</conversation>

The conversation ends with the assistant's answer. The user has not replied yet. Your task: predict the user's next message, assuming it asks for a revision of that answer. Give FIVE distinct plausible requests, most likely first.

`{format_instructions}`
\end{lstlisting}
\end{tcolorbox}
\caption{
Prompts for predicting edit request
}
\label{prompt:pred}
\end{figure*}

\begin{figure*}[tp]
\begin{tcolorbox}[promptboxjudge, boxsep=0pt, left=0pt, right=0pt, top=5.8pt, bottom=5.8pt, lefttitle=6.8pt, righttitle=6.8pt, toptitle=4.8pt, bottomtitle=4.8pt, title={Judge (stage 1): System prompt}]
\begin{lstlisting}[style=promptstyle, xleftmargin=6.8pt, xrightmargin=6.8pt]
You are an AI assistant that decomposes an edit request into a set of requirements. These requirements will later be used to compare two edit requests and determine whether they are essentially equivalent.

You will receive the preceding context (conversations between a user and an LLM), the latest assistant's answer, and an edit request for the answer. You will then break the edit request into its requirements: atomic things the user asks for, each with its target (which part or aspect of the answer) and its change (what should be different, including the direction and any specifics: which name, which section, which value, which style). For example, "make it shorter" has one requirement (target: the answer; change: make it shorter). "Use Australian addresses, remove the numbering and give me 100 entries" has three: (1) target: the addresses; change: make them Australian; (2) target: the numbering of the entries; change: remove it; (3) target: the number of entries; change: make it 100. Also say whether the edit request brings in information that is absent from both the preceding context and the answer (new names, facts, constraints, goals, topics) and list those items.

Write each requirement only from what the edit request itself says. Do not add how the assistant should carry it out, a fix or method the user did not name, or details, examples or specifics the user did not give. The edit request must satisfy every requirement you write. For example, "it throws an IndexError" asks to fix that error (one requirement), not for a particular fix; "more concrete" asks to make the answer more concrete, not to add a project name; a request written in another language that asks for a translation asks for that translation, without inventing its scope beyond what is clear. The language the edit request is written in is not a requirement: add a requirement about the language of the answer only when the edit request asks for one (e.g. a translation). When the edit request can be read in more than one way, choose the reading that fits the answer and the preceding context.
\end{lstlisting}
\par\vspace{4pt}\noindent{\fboxsep=2.8pt\colorbox{orange!30}{\makebox[\dimexpr\linewidth-5.6pt][l]{\hspace*{4.8pt}\ttfamily\bfseries\footnotesize\strut User prompt}}}\par\vspace{3pt}
\begin{lstlisting}[style=promptstyle, xleftmargin=6.8pt, xrightmargin=6.8pt]
The preceding context:
<<<
`{context}`
>>>

The assistant's answer:
<<<
`{artifact}`
>>>

The edit request:
<<<
`{actual}`
>>>

Please decompose the above edit request into a set of requirements.
`{format_instructions}`
\end{lstlisting}
\end{tcolorbox}
\caption{Prompts for decomposing an edit request into requirements.
}
\label{prompt:judge-reqs}
\end{figure*}

\begin{figure*}[tp]
\begin{tcolorbox}[promptboxjudge, boxsep=0pt, left=0pt, right=0pt, top=5.8pt, bottom=5.8pt, lefttitle=6.8pt, righttitle=6.8pt, toptitle=4.8pt, bottomtitle=4.8pt, title={Judge (stage 2): System prompt}]
\begin{lstlisting}[style=promptstyle, xleftmargin=6.8pt, xrightmargin=6.8pt]
You are an AI assistant that judges whether predicted edit requests ask for the same edit as an actual edit request. The actual edit request has already been decomposed into a set of requirements.

You will receive the preceding context (conversations between a user and an LLM), the latest assistant's answer, the actual edit request for the answer, its requirements, and several predicted edit requests written by a system that had not seen the actual one. For each predicted edit request and each requirement, decide:
  satisfied   - the predicted edit request explicitly asks for this change: same target, same direction, same specifics. A more specific version counts (e.g. "cut the opening paragraph" satisfies "make it shorter"). A broader instruction that does not state the change does NOT count ("rewrite it to be better" does not satisfy "make it shorter"); do not infer satisfaction from what a rewrite would probably produce.
  contradicted - the predicted edit request asks for the opposite or for something incompatible.
  absent      - neither.
Extra things the predicted edit request asks for are ignored unless they contradict a requirement. Quote the decisive words of the predicted edit request for each satisfied or contradicted requirement.

The requirements restate what the actual edit request asks for. A predicted edit request that asks for it the way the actual edit request does also satisfies it: one that reports the same error or problem asks for it to be fixed, and one that states the same complaint or need asks for the corresponding change. The language a predicted edit request is written in does not matter: one written in another language that asks for the same change satisfies the requirement. A requirement to keep something as it is, or as it was before, is satisfied by a predicted edit request that does not ask to change it. Judge the request texts; the preceding context and the answer are given only so you can resolve references like "it", "the second paragraph" or "the function".
\end{lstlisting}
\par\vspace{4pt}\noindent{\fboxsep=2.8pt\colorbox{orange!30}{\makebox[\dimexpr\linewidth-5.6pt][l]{\hspace*{4.8pt}\ttfamily\bfseries\footnotesize\strut User prompt}}}\par\vspace{3pt}
\begin{lstlisting}[style=promptstyle, xleftmargin=6.8pt, xrightmargin=6.8pt]
The preceding context:
<<<
`{context}`
>>>

The assistant's answer:
<<<
`{artifact}`
>>>

The actual edit request:
<<<
`{actual}`
>>>

The requirements of the actual edit request:
`{requirements}`

The predicted edit requests:
`{preds}`

Please judge each predicted edit request against every requirement. Use the requirements exactly as given; do not add, merge or split them.
`{format_instructions}`
\end{lstlisting}
\end{tcolorbox}
\caption{Prompts for judging predicted edit requests against the requirements.
}
\label{prompt:judge-verdicts}
\end{figure*}

\begin{figure*}[tp]
\begin{tcolorbox}[promptboxjudge, boxsep=0pt, left=0pt, right=0pt, top=5.8pt, bottom=5.8pt, lefttitle=6.8pt, righttitle=6.8pt, toptitle=4.8pt, bottomtitle=4.8pt, title={Judge (stage 1, check): Rewording prompt}]
\begin{lstlisting}[style=promptstyle, xleftmargin=6.8pt, xrightmargin=6.8pt]
Rewrite the following message in different words, keeping exactly the same meaning and level of detail. Reply with the rewritten message only.

<<<
`{text}`
>>>
\end{lstlisting}
\par\vspace{4pt}\noindent{\fboxsep=2.8pt\colorbox{orange!30}{\makebox[\dimexpr\linewidth-5.6pt][l]{\hspace*{4.8pt}\ttfamily\bfseries\footnotesize\strut Rewrite prompt}}}\par\vspace{3pt}
\begin{lstlisting}[style=promptstyle, xleftmargin=6.8pt, xrightmargin=6.8pt]
A rewording of the edit request, with the same meaning, was checked against your requirements:
<<<
`{rephrased}`
>>>

It did not satisfy these requirements:
`{failures}`

Rewrite the requirements so that any request with the same meaning as the edit request, including this rewording, satisfies every one of them. Follow the same rules: write each requirement only from what the edit request itself says, and keep every specific it states (names, values, places, styles). If the rewording does not mean the same as the edit request, do not follow it: keep the requirements that the edit request itself supports. Return the full set of requirements again.
`{format_instructions}`
\end{lstlisting}
\end{tcolorbox}
\caption{Prompts for checking the requirements with a rewording of the edit request.
}
\label{prompt:judge-check}
\end{figure*}

\begin{figure*}[tp]
\begin{tcolorbox}[promptboxgate, boxsep=0pt, left=0pt, right=0pt, top=5.8pt, bottom=5.8pt, lefttitle=6.8pt, righttitle=6.8pt, toptitle=4.8pt, bottomtitle=4.8pt, title={Decision model: state}]
\begin{lstlisting}[style=promptstyle, xleftmargin=6.8pt, xrightmargin=6.8pt]
{"conversation": [{"role": "user", "content": `{turn_1}`},
                  {"role": "assistant", "content": `{turn_2}`}, ...],
 "output_to_revise": `{artifact}`,
 "suggestions": [`{suggestion_1}`, ..., `{suggestion_5}`]}
\end{lstlisting}
\par\vspace{4pt}\noindent{\fboxsep=2.8pt\colorbox{cyan!32}{\makebox[\dimexpr\linewidth-5.6pt][l]{\hspace*{4.8pt}\ttfamily\bfseries\footnotesize\strut Decision model: questions}}}\par\vspace{3pt}
\begin{lstlisting}[style=promptstyle, xleftmargin=6.8pt, xrightmargin=6.8pt]
match: Will at least one of the suggestions explicitly ask for every change (same target, direction and specifics) that the user's next request to revise output_to_revise will ask for, and for nothing incompatible with it? A broader instruction that does not state a change does not count.

s`{i}`: Will suggestion `{i}` (item `{i}` of suggestions) explicitly ask for every change (same target, direction and specifics) that the user's next request to revise output_to_revise will ask for, and for nothing incompatible with it? A broader instruction that does not state a change does not count.
\end{lstlisting}
\end{tcolorbox}
\caption{Input state and questions of the decision model.
}
\label{prompt:gate}
\end{figure*}

\end{document}